\documentclass[11pt]{article}

\usepackage[final]{acl}

\usepackage{times}
\usepackage{latexsym}
\usepackage{tcolorbox}
\usepackage{subcaption}
\usepackage{caption}
\usepackage{tikz}
\usetikzlibrary{positioning}
\usepackage[ragged]{footmisc}
\usepackage{hyperref}
\usepackage{booktabs}

\usepackage{tikz}
\usetikzlibrary{arrows.meta, calc, positioning}
\usepackage{amsmath, amssymb}

\usepackage{amsmath,amssymb,amsfonts}
\usepackage{algorithmic}
\usepackage{graphicx}
\usepackage{textcomp}
\usepackage{xcolor}
\usepackage{enumitem}
\usepackage{multirow}
\usepackage{xurl}
\usepackage{hyperref}
\usepackage{stfloats}

\usepackage[T1]{fontenc}
\usepackage[utf8]{inputenc}
\usepackage{microtype}
\usepackage{inconsolata}
\usepackage{graphicx}

\title{Judging LLM-as-a-Judge:
Concerning Rubric Artifacts in LLM-based Automated Text Generation Evaluation}

\author{
  \textbf{Anshul Bagaria},
  \textbf{Sowmya S Sundaram},
  \textbf{Gokul S Krishnan},
  \textbf{Balaraman Ravindran}, \\
  Centre for Responsible AI (CeRAI), Wadhwani School of Data Science and AI (WSAI),\\ IIT Madras, Chennai, India\\
  \texttt{be21b005@smail.iitm.ac.in},
  \texttt{\{sowmya,gokul\}@cerai.in},
  \texttt{ravi@dsai.iitm.ac.in}
}

\begin{document}
\maketitle

\begin{abstract}
LLM-as-a-Judge pipelines are increasingly used to evaluate AI-generated text, based on the assumption that judgments arise from reasoning over candidate responses with respect to a rubric. We show that this assumption warrants further scrutiny. Classifiers trained only on rubric text, without access to any evaluated response, achieve nontrivial predictive performance on judge outputs. This suggests that rubric formulations encode recoverable evaluative signals, allowing scores to be partially anticipated independently of model outputs. Finally, counterfactual perturbations reveal that judges often fail to reliably update their decisions when either the candidate response or the rubric criterion is reversed. Our findings raise concerns about the reliability of rubric-based LLM evaluation and highlight the need for further methodological study of automated evaluation via LLMs.
\end{abstract}

\section{Introduction}
LLM-based evaluators are now widely used to benchmark and validate generative systems, despite growing evidence of systematic biases such as position effects, verbosity preferences, and sensitivity to prompting~\cite{shi2025judging, chen2024humans, li2026evaluating, gu2026survey, tan2025judgebench}. These issues raise a broader concern: whether current evaluation pipelines reliably measure response quality or reflect artifacts of the evaluation setup.

\begin{figure*}
\centering
    \includegraphics[scale=0.3]{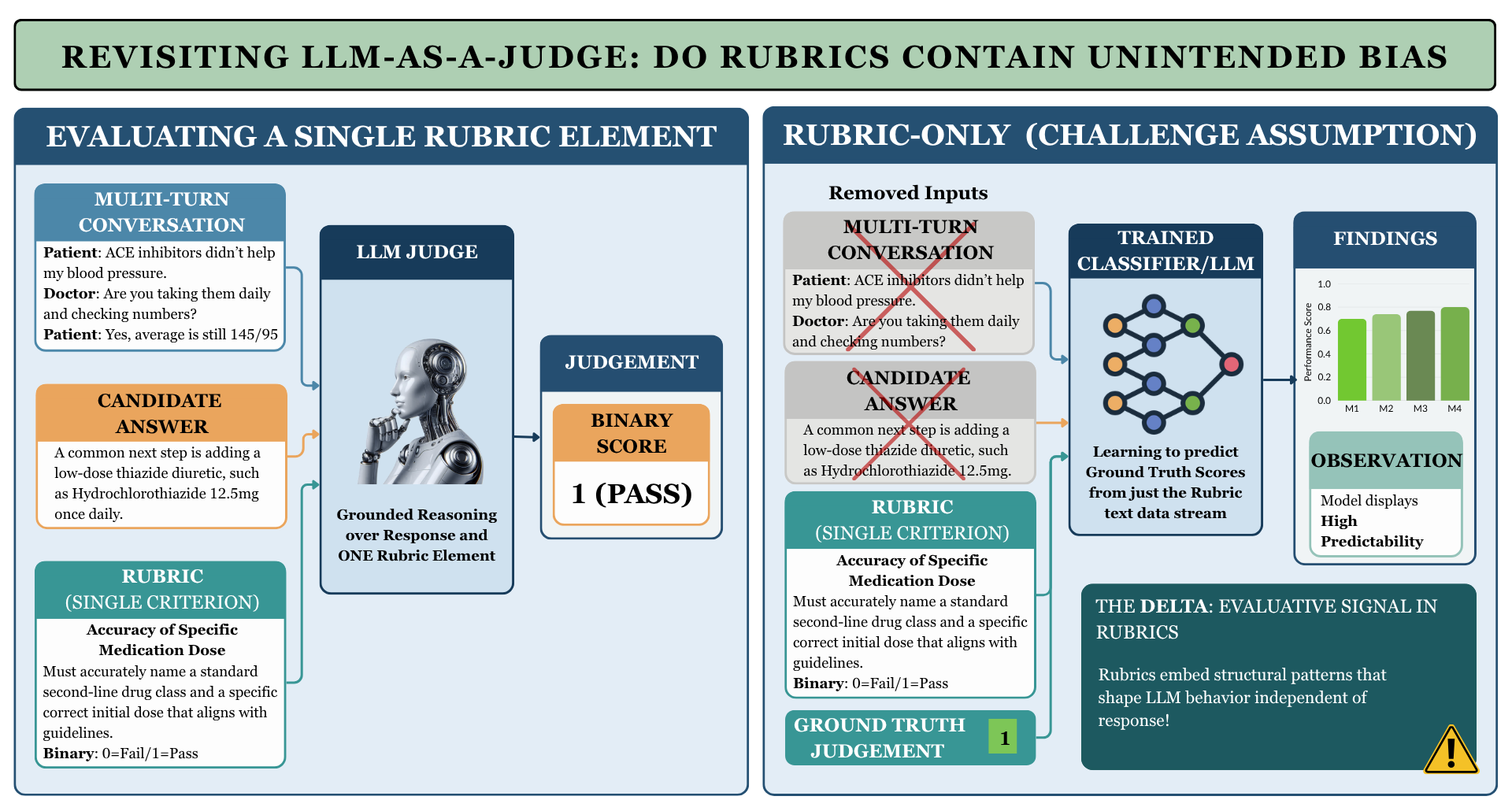}
    \caption{\textbf{Rubric Probe}: Given a rubric alone, we train a classifier to predict the judge label, testing whether rubric text carries recoverable evaluative signal beyond chance.}
    \label{fig:arch}
\end{figure*}

\noindent Building on critiques of LLMs as judges, there are many evaluation settings, including direct scoring, pairwise comparison, and reference-based scoring, which vary in the structure of inputs and the guidance provided~\cite{gu2026survey}. Among these, rubric-based evaluation is widely adopted, especially in high-stakes domains, because it provides explicit criteria intended to standardize judgments and improve transparency~\cite{croxford2025evaluating}. This reliance rests on the assumption that rubrics constrain models without introducing an independent evaluative signal. We test this assumption by proposing a novel controlled experiment (Figure~\ref{fig:arch}) that asks whether an evaluative signal can be recovered from the rubric alone, without access to the candidate response or its context. This formulation isolates the rubric's contribution and removes systemic biases that arise in response evaluation.

\noindent Using \textsc{HealthBench} \cite{arora2025healthbench} and \textsc{ResearchRubrics} \cite{sharma2026researchrubrics}, we find that rubric text alone predicts judge outputs, suggesting a previously underexamined coupling between rubric formulation and evaluation outcomes. Our analysis of embedding geometry suggests that rubric text conveys an evaluative signal. This motivates closer scrutiny of the rubric design in LLM-based evaluation.

\section{Related Work}
Prior work on LLM-as-a-Judge has shown that automated evaluation is sensitive to position effects, scoring bias, prompt framing, and other non-semantic factors, raising concerns about the reliability of model-based judgments~\cite{zheng2023judging, shi2025judging, chen2024humans, li2026evaluating, tan2025judgebench, gu2026survey}. In benchmark settings such as \textsc{HealthBench}, where expert-authored rubrics define the evaluation criteria for medical responses, and \textsc{ResearchRubrics}, with instance-specific rubrics, rubrics are typically treated as evaluation instructions rather than potential sources of signal ~\cite{arora2025healthbench, sharma2026researchrubrics}. Our work differs in that it probes whether the rubric text itself encodes recoverable evaluative priors, independent of the candidate response. This connects to shortcut-learning research showing that language models often exploit superficial lexical cues rather than intended semantic signal~\cite{shortcutmaze2024, ong2024shortcut}. 

\section{LLM-as-a-Judge Probe Design}
\begin{figure}
\centering
\resizebox{\columnwidth}{!}{%
\begin{tikzpicture}[
    font=\small,
    node distance=0.25cm and 0.25cm,
    box/.style={draw, rounded corners=2pt, very thick, minimum height=0.95cm, align=center, inner sep=6pt},
    rubric/.style={box, fill=blue!12, draw=blue!60!black},
    probe/.style={box, fill=purple!14, draw=purple!70!black, minimum width=3.0cm},
    label/.style={box, fill=green!12, draw=green!55!black},
    hidden/.style={draw, rounded corners=2pt, dashed, thick, gray!55, fill=gray!8, text=gray!65, minimum height=0.75cm, align=center, inner sep=5pt},
    arrow/.style={->, very thick, draw=black!75},
    dashedlink/.style={->, dashed, thick, draw=orange!75!black, opacity=0.9, >=Stealth},
    note/.style={font=\small, text=black}
]

\node[rubric] (rubric) {Rubric text\\ \(r\)};
\node[probe, right=of rubric] (probe) {Probe classifier};
\node[label, right=of probe] (label) {LLM-as-a-judge\\ output \(y\)};

\draw[arrow] (rubric) -- (probe);
\draw[arrow] (probe) -- (label);

\node[hidden, below=1.0cm of probe] (answer) {LLM answer};
\node[hidden, left=of answer] (meta) {LLM Query};
\node[hidden, right=of answer] (ground) {Grounding\\ information};

\draw[dashedlink] (answer.north) to[bend left=18] (label.south);
\draw[dashedlink] (meta.north) to[bend right=10] (label.south);
\draw[dashedlink] (ground.north) to[bend left=10] (label.south);
\draw[dashedlink] (rubric.south) to[bend right=10] (label.south);

\node[note, above=0.15cm of label] {Chance level: 0.5};

\end{tikzpicture}%
}

\caption{\textbf{Probe overview}. The classifier receives only rubric text and predicts the LLM-as-a-judge output, while the evaluated answer, contextual metadata, and grounding information are excluded.}
\label{fig:probe}
\end{figure}
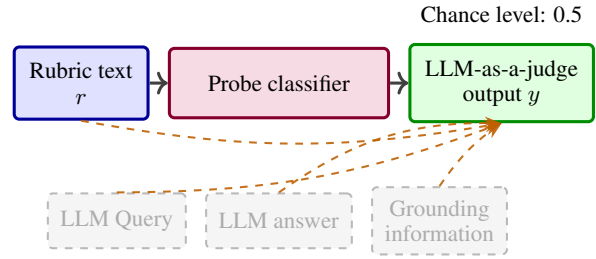

A central assumption in rubric-based evaluation is that judgments arise from grounded reasoning over the candidate response with respect to the rubric, which is treated as an evaluative instruction. We instead investigate whether rubric formulations encode latent priors correlated with evaluation labels, such that a model can partially infer the expected judgment from the rubric text alone (Figure~\ref{fig:probe}).

\begin{tcolorbox}[colback=gray!5,colframe=gray!75!black,title=Formulation]
\small
Let \(r\) denote a rubric, \(z\) the input context, \(x\) a candidate response, and \(y\) the judge label. LLM-as-a-Judge aims to model
\[
p(y \mid r, z, x),
\]
where the rubric conditions how responses are evaluated. For evaluation to be valid, the rubric should guide judgment rather than independently determine the label. In particular, \(y\) should not be predictable from \(r\) alone:
\[
p(y \mid r) \approx 0.5.
\]

We empirically test whether
\[
p(y \mid r) > 0.5,
\]
which indicates that the rubric text contains a standalone predictive signal about \(y\) that is independent of the response \(x\) and the context \(z\).
\end{tcolorbox}

\noindent Beyond predictive performance, we analyze the structure of the rubric signal using generalizability studies and the geometry of the embedding space.

\section{LLM-as-a-Judge Probe Experiment}
We start by describing the details of our experiment and setup, followed by the probe results.

\subsection{Experiment Details}
We evaluate our probe across two complementary benchmarks:
\textsc{HealthBench}, including its \textsc{HealthBench-Eval} and
\textsc{HealthBench-Hard} variants, which provide conversation-specific rubrics for open-ended healthcare responses, and \textsc{ResearchRubrics}, which use instance-specific rubrics to assess whether research queries are sufficiently answered.
\begin{figure*}[t]
    \includegraphics[width=\textwidth]{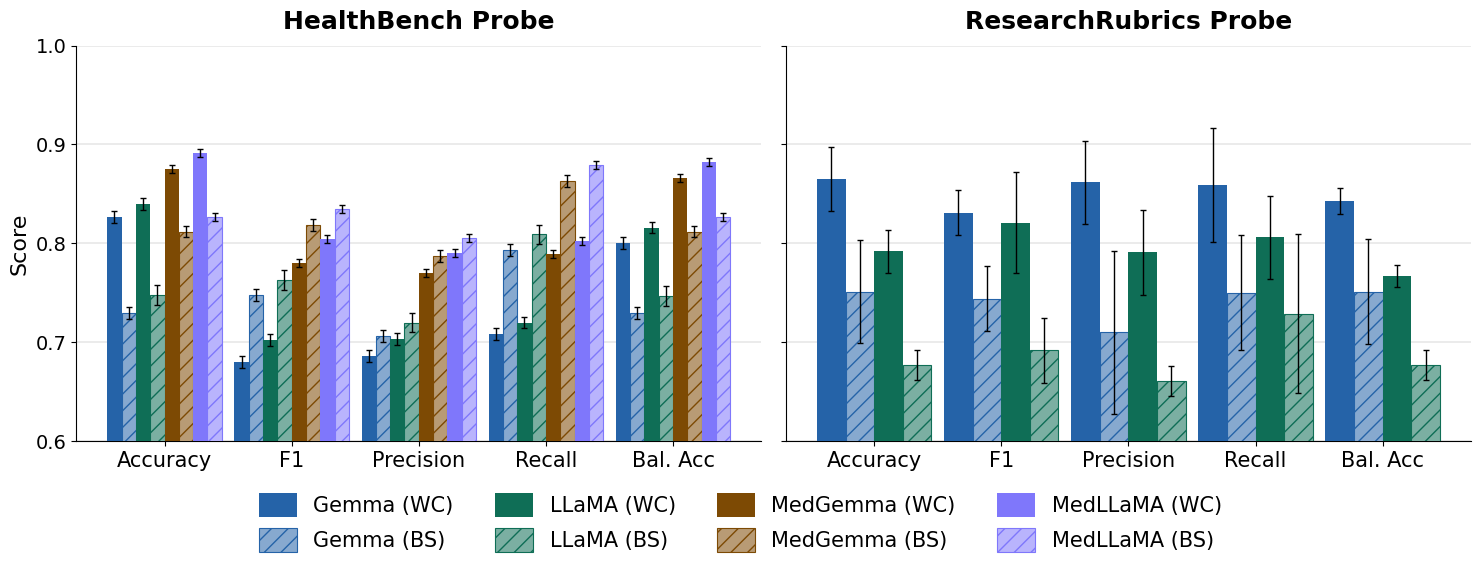}
    \caption{\textbf{Per-metric comparison across probe datasets.} PubMedBERT classifiers use rubric text to predict LLM-as-a-Judge labels. The left panel shows \textsc{HealthBench} results for Gemma, LLaMA, MedGemma, and MedLLaMA, while the right shows \textsc{ResearchRubrics} results for Gemma and LLaMA. The latter reflects the available candidate-model evaluations for this non-medical domain. Error bars denote standard deviation. The weighted classification (WC) and balanced subsampling (BS) cohorts are presented here to address class imbalance.}
    \label{fig:main-results}
\end{figure*}

\noindent\textbf{Dataset Source:} We analyze two complementary benchmarks with instance-specific rubrics: \textsc{HealthBench} and \textsc{ResearchRubrics}. \textsc{HealthBench} \cite{arora2025healthbench} is a comprehensive healthcare dataset comprising complex healthcare conversations, where each conversation is paired with a set of case-specific rubrics. More importantly, the rubrics are provided by clinicians and are NOT generated by an LLM. These features, which follow best-practice benchmarking, made \textsc{HealthBench} a strong candidate for the analysis. In the standard evaluation setting, an LLM generates a response that is evaluated against the corresponding rubric by an LLM-as-a-Judge, producing a binary satisfiability label (0/1). We have 5k such conversations paired with an expected answer and case-specific rubrics.

\noindent\textsc{ResearchRubrics} \cite{sharma2026researchrubrics} provides a complementary non-medical setting, pairing research queries with instance-specific rubrics that assess whether the query has been sufficiently answered. Thus, while the two benchmarks differ in domain and task, both employ per-instance rubrics, allowing us to examine whether rubric-derived evaluative signals persist across distinct evaluation settings.

\noindent\textbf{Probing Dataset:} We first run the candidate LLM and obtain responses from various models. We then apply the LLM-as-a-Judge model and obtain the class labels (0/1). We use Qwen (Qwen2.5-7B-Instruct~\cite{qwen2025qwen25technicalreport}) as an LLM-as-a-Judge on LLM outputs from LlaMA (meta-llama/Llama-3.1-8B-Instruct~\cite{grattafiori2024llama}), MedLlaMA (MMed-Llama-3-8B~\cite{qiu2024towards}), Gemma (google/gemma-7b-it~\cite{team2024gemma}), and MedGemma (google/medgemma-1.5-4b-it~\cite{sellergren2026medgemma}). Then, we construct four classifier datasets, using the rubric as input and the LLM-as-a-Judge output as the label for each model. \textsc{HealthBench-Eval-Probe} comprises approximately $\approx$43k rubrics, while \textsc{HealthBench-Hard-Probe} contains around $\approx$11k rubrics. The final experiment evaluates eight cohorts across these two datasets. Additionally, \textsc{ResearchRubrics-Probe} contains approximately $\approx$2.5k rubrics and is used to assess generalization beyond the healthcare domain. Each classifier is trained using an 80/20 train–test split.

\noindent\textbf{Classifier Setup:} We use the PubMedBERT~\cite{gu2021domain} based classifier for our main experiments. For generalization, we test cross-dataset transfer between \textsc{HealthBench-Eval-Probe} and \textsc{HealthBench-Hard-Probe} by evaluating on \textsc{ResearchRubrics-Probe}. \footnote{Code available here: \href{https://anonymous.4open.science/r/JudgingLLM-as-a-Judge-CFA6/}{Repository Link}}
\subsection{Rubric Probe Results}
In this study, we evaluate the PubMedBERT-based classifier's performance against the four cohorts (Gemma, MedGemma, LlaMA, MedLLaMa), across the \textsc{HealthBench-Eval-Probe} and \textsc{HealthBench-Hard-Probe} using standard metrics while explicitly accounting for class imbalance through weighted classification and balanced subsampling strategies (Figure~\ref{fig:main-results}). We observe that the classifier outputs are \textbf{> 0.5 and, in some cases, exceeding 0.8}, even after accounting for dataset imbalance. A similar pattern is observed on \textsc{ResearchRubrics}, suggesting that rubric text carries predictive signal beyond the healthcare setting. Together, these results indicate that the rubric text contains a signal that predicts judge labels beyond random chance. Additional analyses on these classifier experiments are included in the Appendix~\ref{extra-classifier}. 

\subsection{Generalization Checks}
We perform several checks to assess the robustness and generalizability of the rubric-only predictive signal. First, we evaluate the PubMedBERT-based probe using 5-fold cross-validation and bootstrap confidence intervals, rather than relying on a single train–test split. Detailed results from these evaluations are reported in Appendix~\ref{app:cross_val_proto}.

\noindent We additionally examine whether the observed performance is sensitive to the choice of classifier architecture. We compare PubMedBERT with BERT, RoBERTa, and DeBERTa-v3, and report the corresponding results alongside majority-class baselines in Appendix~\ref{app:classifier-ablations}. These comparisons help distinguish the observed signal from performance attributable to a particular classifier or class imbalance.

\noindent Finally, we conduct additional transfer and diagnostic analyses, including cross-dataset evaluation between HEALTHBENCH-EVAL-PROBE and HEALTHBENCH-HARD-PROBE and experiments using simpler classifiers such as TF-IDF with logistic regression. These results are provided in Appendix~\ref{app:cross_val_proto}. Collectively, these analyses provide complementary checks on the robustness of the rubric-only predictive signal.

\section{Counterfactual Consistency Analysis}
\label{sec:counterfactual}
The rubric-only probe establishes an association between rubric text and judge decisions, but this association could in principle arise from differences in item difficulty rather than rubric-conditioned evaluation. To distinguish these possibilities, we conduct \textbf{two paired counterfactual experiments} that directly test whether the judge responds to controlled changes in either the candidate response or the evaluation criterion. Detailed experimental setup and results are provided in Appendix~\ref{app:counterfactual}.

\begin{figure}[t]
    \centering
    \includegraphics[width=0.8\linewidth]{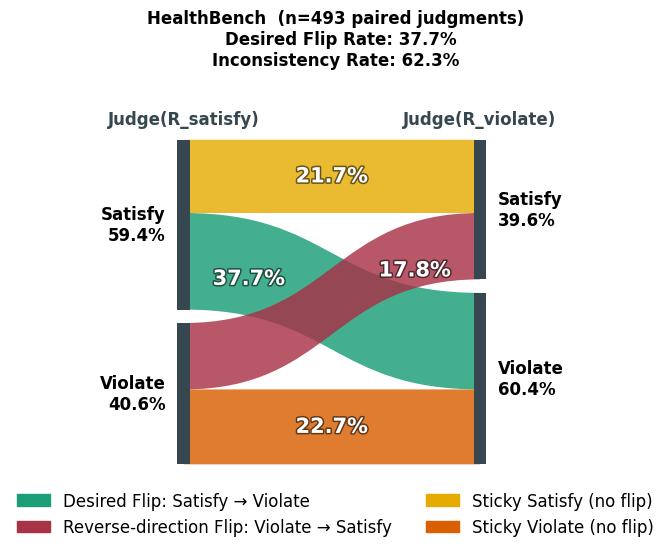}
    \caption{\textbf{Output perturbation}: The conversation and rubric are fixed while the candidate response is reversed. A counterfactually responsive judge should always flip its verdict. Instead, the expected verdict reversal occurs in only 37.7\% of pairs, indicating that the judge often fails to track changes in answer-level rubric satisfaction.}
    \label{fig:output_counterfactual}
\end{figure}

\subsection{Output Perturbation}
\label{sec:output-perturbation}
We first test whether the judge responds to changes in the candidate's response while holding the conversation and rubric fixed. We randomly sample 500 \textsc{HealthBench} conversations and select one rubric per conversation. For each pair, Mistral-7B generates responses designed to respectively satisfy and violate the rubric, followed by a qualitative check. Qwen2.5-7B-Instruct then evaluates both responses under the same context and rubric.

\noindent A counterfactually responsive judge should reverse its verdict when response-level evidence is reversed. Across 493 valid pairs, however, the judge produces the \textbf{expected verdict reversal in only 37.7\% of cases}, yielding an inconsistency rate of 62.3\% (Figure~\ref{fig:output_counterfactual}). Since the context and rubric are fixed within each pair, these failures cannot be attributed solely to cross-item differences in difficulty or rubric wording. While rubric ambiguity or response-construction errors may contribute, the results demonstrate substantial non-responsiveness to controlled changes in answer-level evidence.

\subsection{Rubric Perturbation}
\label{sec:rubric-counterfactual}
We next test whether the judge tracks changes in the evaluation criterion. We randomly sample 1,000 \textsc{HealthBench} conversations with candidate responses from Gemma and LLaMA, and construct counterfactual rubrics whose intended criteria are reversed using Mistral-7B. The conversation and candidate response remain fixed, so only the rubric semantics change. We again perform a manual qualitative check of the paired rubrics.

\begin{figure}[t]
    \centering
    \includegraphics[width=\linewidth]{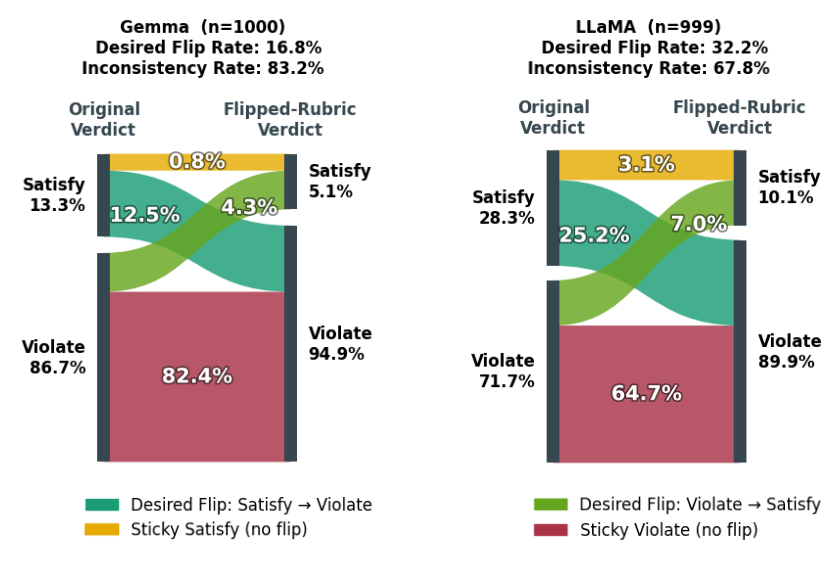}
    \caption{\textbf{Rubric perturbation}: On \textsc{HealthBench}, the conversation and candidate response are fixed while the rubric criterion is reversed. A criterion-sensitive judge should therefore always flip its verdict. Instead, the expected reversal occurs in only 16.8\% of pairs with Gemma and 32.2\% of pairs with LLaMA, indicating poor sensitivity to changes in the evaluation criterion.}
    \label{fig:rubric_perturbation}
\end{figure}

\noindent A criterion-sensitive judge should reverse its verdict when the rubric meaning is reversed. Instead, the desired flip rate is only \textbf{16.8\% for Gemma and 32.2\% for LLaMA}, corresponding to inconsistency rates of 83.2\% and 67.8\%, respectively (Figure~\ref{fig:rubric_perturbation}). Thus, although rubric changes can affect judgments, the judge retains its original verdict despite a reversal of the evaluation criterion.

\begin{figure}[b!]
    \centering
    \includegraphics[width=0.7\columnwidth]{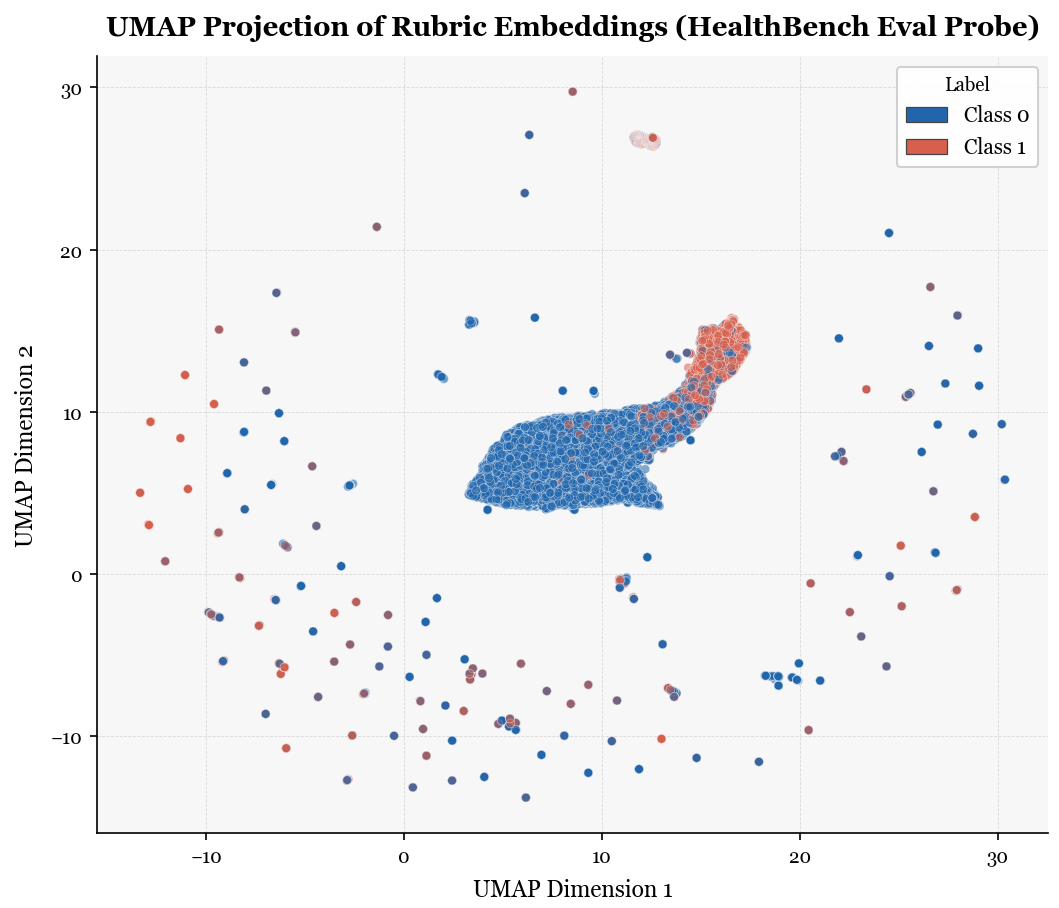}
    \includegraphics[width=0.7\columnwidth]{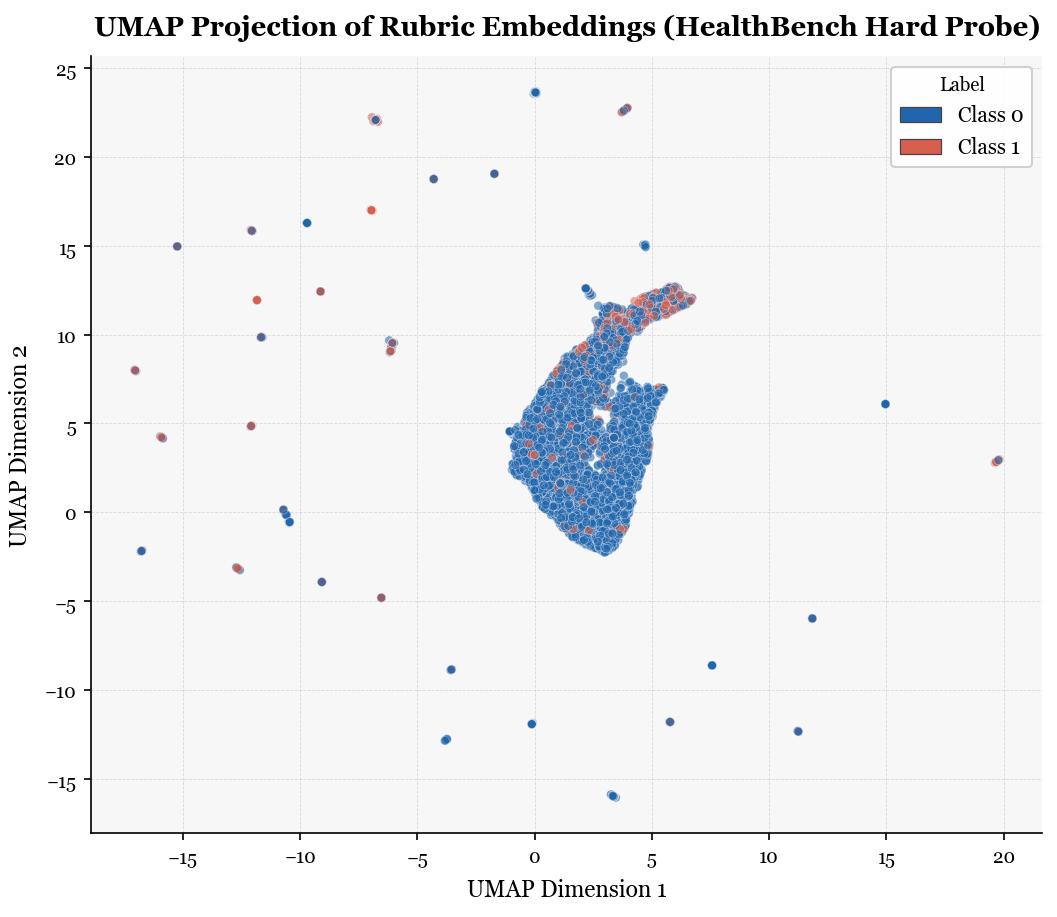}
    \caption{The UMAP projections of \textsc{HealthBench-Eval-Probe} and \textsc{HealthBench-Hard-Probe} }
    \label{fig:umap}
\end{figure}


\subsection{Semantic Analysis of Rubrics}
We next probe whether the rubrics themselves, independent of the LLM-as-a-Judge output, encode semantic cues that may drive the observed label structure. With UMAP~\cite{mcinnes2018umap}, we visualize BERT embeddings to test whether Eval and Hard rubrics form separable clusters (Figure~\ref{fig:umap}). \textsc{HealthBench-Hard} contains more challenging and clinically ambiguous cases than \textsc{HealthBench-Eval}, providing a useful contrast in rubric characteristics. Additional semantics are included in Appendix~\ref{app:semantics}.

\noindent The correct label is defined by the rubric, the context, and the candidate response, and should not be recoverable from the rubric alone. Yet we observe consistent correlations between rubric text and judge labels even without the response. Moreover, \textsc{HealthBench-Eval} rubrics are more separable with respect to judge labels, suggesting that some rubric formulations are associated with particular outcomes. Together, these findings indicate that rubric text encodes label-correlated signal, raising concerns that evaluation may reflect rubric properties independently of the response. 

\subsection{Discussion}
Our probe shows that rubric text alone predicts LLM-as-a-Judge outputs above chance, demonstrating a recoverable rubric-conditioned signal independent of the candidate response. This signal is distributed across the semantic structure rather than being reducible to keyword artifacts. These patterns indicate that evaluation difficulty is driven not solely by the task but also by variability and ambiguity in rubric formulation. This is not a model capacity limitation. Our findings are unlikely to be explained by dataset-specific confounding alone, since the probe never observes the response or context, and the effect persists across \textsc{HealthBench-Probe} splits and the \textsc{ResearchRubrics} setting. While we do not establish causality, we identify a consistent association between rubric text and judge outputs. This motivates systematic validation of rubric design, including stress-testing rubric sensitivity, auditing judge rationales~\cite{delucia2026same}, and complementing rubric-based evaluation with more robust metrics~\cite{shailya2025lext, ito2025reference, croxford2025evaluating}.

\section{Conclusion}
Rubric-conditioned evaluation may inadvertently encode predictive signal that correlates with judges' outputs. To make LLM-as-a-Judge reliable, benchmarks should aim to eliminate recoverable rubric priors. We hope this work motivates closer scrutiny of rubric design in automated evaluation.

\section{Limitations}
Our probe intentionally isolates rubric text, thereby capturing potential rubric priors rather than end-to-end judge behavior in full evaluation settings. We evaluate \textsc{HealthBench} and \textsc{ResearchRubrics}, but the findings may not generalize to other domains, rubric styles, or languages. We also rely on a small set of probing methods and a binary formulation, which may understate or distort more subtle rubric effects. Finally, while the results are consistent with shortcut learning, they do not identify the full causal mechanism behind rubric-conditioned judgments. We posit that, despite these limitations, our work serves as an initial probe to invite research into the design of automated text evaluation.

\section*{AI Usage Declaration}
We acknowledge the use of AI for writing assistance, including grammatical corrections and minor rewrites.
\bibliography{custom}

\appendix
\section{Experimental Setup and Modeling}
\label{app}

\subsection{Models Explored}
\label{app_1}
We evaluate four candidate language models for generating responses to the \textsc{HealthBench} conversations. The models are governed by distinct open licenses that permit research use: \textit{Llama-3.1-8B-Instruct} is licensed under the \href{https://llama.meta.com/llama3_1/license/}{\textbf{Llama 3.1 Community License Agreement}}, \textit{Gemma-7b-it} is governed by the \href{https://ai.google.dev/gemma/terms}{\textbf{Gemma Terms of Use}}, and \textit{MedGemma} model is licensed under the \href{https://developers.google.com/health-ai-developer-foundations/terms}{\textbf{Health AI Developer Foundations License}} but the repository is licensed under the  \href{https://www.apache.org/licenses/LICENSE-2.0}{\textbf{Apache 2.0 License}}. \textit{MedLLaMA} is licensed under \href{https://creativecommons.org/licenses/by-nc-nd/4.0/deed.en}{\textbf{CC-BY-NC-ND}}. Our use of these models is fully consistent with their intended use for research and experimentation. The evaluation of model biases and limitations aligns with the responsible AI development practices encouraged by their creators and complies with the \href{https://llama.meta.com/llama3_1/use-policy/}{\textbf{Llama 3.1 Acceptable Use Policy}}, the \href{https://ai.google.dev/gemma/prohibited_use_policy}{\textbf{Gemma Prohibited Use Policy}}, the \href{https://huggingface.co/johnsnowlabs}{\textbf{MedLlaMA Acceptable Use Policy} and the \href{https://developers.google.com/health-ai-developer-foundations/terms}{\textbf{Health AI Developer Foundations Terms of Use}}}.

\subsection{Computing Requirements}
The experimental pipeline was implemented in Python 3.10. Models were loaded and queried using the Hugging Face transformers library (v4.38.2) with the PyTorch (v2.1) backend. All the experiments were executed on a GPU cloud instance with NVIDIA A100 GPUs to accelerate inference. Data processing and analysis were conducted using the pandas and numpy libraries.

\noindent A total of 5,000 Question Answering operations, along with 43,000 Yes/No answers for LLM-as-a-judge answers across 4 model generations, were performed. The total computational budget is estimated at 30-35 GPU-hours on the specified hardware.

\noindent This study evaluates pre-trained models, so no model training or fine-tuning was performed. The key hyperparameters relate to the text generation (decoding) process. To ensure a fair and consistent comparison across all models, a fixed set of decoding parameters was used for every query detailed in \autoref{tab:hyperparams}. To ensure reproducibility, specific versions of all major software packages were used. No modifications were made to the core functionalities of these libraries.

\textbf{Core ML/DL Libraries}: transformers (v4.38.2), torch (v2.1).

\textbf{Data Handling Tools}: pandas (v2.0.3), numpy (v1.25.2).
\begin{table}[h!]
  \centering
  \begin{tabular}{lll}
    \toprule
    \textbf{Parameter} & \textbf{Value} \\
    \midrule
    temperature          & 1.0 \\
    repetition penalty & 1.15 \\
    top\_p                & 0.95 \\
    num\_return\_sequences & 1     \\
    \bottomrule
  \end{tabular}
  \caption{Decoding hyperparameters used for all model queries. All the other parameters were set to default values.}
  \label{tab:hyperparams}
\end{table}

\section{Rubric Classification Exploration}
\label{extra-classifier}
Apart from the main classification results, we now present deeper analysis in precision and recall through confusion matrices for an illustration (Figures~\ref{fig:gemma_cm}, ~\ref{fig:llama_cm}, ~\ref{fig:medgemma_cm}, ~\ref{fig:medllama_cm}). We evaluated models on the OpenAI \textsc{HealthBench} dataset, which is released under the \href{https://opensource.org/license/mit}{\textbf{MIT license}} with strict safeguards under \href{https://openai.com/policies/usage-policies/}{OpenAI Usage Policies} and standard \href{https://huggingface.co/content-policy}{HuggingFace Content Policy}.

\noindent To further probe whether the evaluative signal is recoverable from the rubric text alone, we project the rubric inputs into a TF-IDF vector space and evaluate two simple classifiers (multinomial Naïve Bayes and L2-regularized logistic regression) alongside a majority-class dummy baseline.

\begin{figure*}[t]
    \centering
    \begin{subfigure}{0.48\linewidth}
        \begin{subfigure}{0.25\linewidth}
            \centering
            \begin{tikzpicture}[
                box/.style={draw, rectangle, minimum width=1cm, minimum height=0.6cm, text width=1cm, align=center},
                correct/.style={box, fill=green!20},
                error/.style={box, fill=red!20}
            ]
            \matrix (conmat) [row sep=.05cm, column sep=.05cm] {
                \node (tp) [correct] {1043}; &
                \node (fn) [error] {904}; \\
                \node (fp) [error] {757}; &
                \node (tn) [correct] {5982}; \\
            };
            \node [left=.2cm of conmat, yshift=0.7cm, rotate=90, font=\bfseries] {Actual};
            \node [above=.05cm of conmat, font=\bfseries] {Predicted};
            \end{tikzpicture}
            {\small HB Eval Probe}
        \end{subfigure}
        \hspace{1.5cm}
        \begin{subfigure}{0.25\linewidth}
            \centering
            \begin{tikzpicture}[
                box/.style={draw, rectangle, minimum width=1cm, minimum height=0.6cm, text width=1cm, align=center},
                correct/.style={box, fill=green!20},
                error/.style={box, fill=red!20}
            ]
            \matrix (conmat) [row sep=.05cm, column sep=.05cm] {
                \node (tp) [correct] {153}; &
                \node (fn) [error] {150}; \\
                \node (fp) [error] {167}; &
                \node (tn) [correct] {1772}; \\
            };
            \node [left=.2cm of conmat, yshift=0.7cm, rotate=90, font=\bfseries] {Actual};
            \node [above=.05cm of conmat, font=\bfseries] {Predicted};
            \end{tikzpicture}
            {\small HB Hard Probe}
        \end{subfigure}
        \caption{Gemma Model}
        \label{fig:gemma_cm}
    \end{subfigure}
    \hfill
    \begin{subfigure}{0.48\linewidth}
        \begin{subfigure}{0.25\linewidth}
            \centering
            \begin{tikzpicture}[
                box/.style={draw, rectangle, minimum width=1cm, minimum height=0.6cm, text width=1cm, align=center},
                correct/.style={box, fill=green!20},
                error/.style={box, fill=red!20}
            ]
            \matrix (conmat) [row sep=.05cm, column sep=.05cm] {
                \node (tp) [correct] {1114}; &
                \node (fn) [error] {833}; \\
                \node (fp) [error] {392}; &
                \node (tn) [correct] {6347}; \\
            };
            \node [left=.2cm of conmat, yshift=0.7cm, rotate=90, font=\bfseries] {Actual};
            \node [above=.05cm of conmat, font=\bfseries] {Predicted};
            \end{tikzpicture}
            {\small HB Eval Probe}
        \end{subfigure}
        \hspace{1.5cm}
        \begin{subfigure}{0.25\linewidth}
            \centering
            \begin{tikzpicture}[
                box/.style={draw, rectangle, minimum width=1cm, minimum height=0.6cm, text width=1cm, align=center},
                correct/.style={box, fill=green!20},
                error/.style={box, fill=red!20}
            ]
            \matrix (conmat) [row sep=.05cm, column sep=.05cm] {
                \node (tp) [correct] {164}; &
                \node (fn) [error] {139}; \\
                \node (fp) [error] {256}; &
                \node (tn) [correct] {1683}; \\
            };
            \node [left=.2cm of conmat, yshift=0.7cm, rotate=90, font=\bfseries] {Actual};
            \node [above=.05cm of conmat, font=\bfseries] {Predicted};
            \end{tikzpicture}
            {\small HB Hard Probe}
        \end{subfigure}
        \caption{LLaMA Model}
        \label{fig:llama_cm}
    \end{subfigure}
    \centering
    \begin{subfigure}{0.48\linewidth}
        \begin{subfigure}{0.25\linewidth}
            \centering
            \begin{tikzpicture}[
                box/.style={draw, rectangle, minimum width=1cm, minimum height=0.6cm, text width=1cm, align=center},
                correct/.style={box, fill=green!20},
                error/.style={box, fill=red!20}
            ]
            \matrix (conmat) [row sep=.05cm, column sep=.05cm] {
                \node (tp) [correct] {1311}; &
                \node (fn) [error] {636}; \\
                \node (fp) [error] {277}; &
                \node (tn) [correct] {6462}; \\
            };
            \node [left=.2cm of conmat, yshift=0.7cm, rotate=90, font=\bfseries] {Actual};
            \node [above=.05cm of conmat, font=\bfseries] {Predicted};
            \end{tikzpicture}
            {\small HB Eval Probe}
        \end{subfigure}
        \hspace{1.5cm}
        \begin{subfigure}{0.25\linewidth}
            \centering
            \begin{tikzpicture}[
                box/.style={draw, rectangle, minimum width=1cm, minimum height=0.6cm, text width=1cm, align=center},
                correct/.style={box, fill=green!20},
                error/.style={box, fill=red!20}
            ]
            \matrix (conmat) [row sep=.05cm, column sep=.05cm] {
                \node (tp) [correct] {215}; &
                \node (fn) [error] {88}; \\
                \node (fp) [error] {188}; &
                \node (tn) [correct] {1751}; \\
            };
            \node [left=.2cm of conmat, yshift=0.7cm, rotate=90, font=\bfseries] {Actual};
            \node [above=.05cm of conmat, font=\bfseries] {Predicted};
            \end{tikzpicture}
            {\centering \small HB Hard Probe}
        \end{subfigure}
        \caption{MedGemma Model}
        \label{fig:medgemma_cm}
    \end{subfigure}
    \hfill
    \begin{subfigure}{0.48\linewidth}
        \begin{subfigure}{0.25\linewidth}
            \centering
            \begin{tikzpicture}[
                box/.style={draw, rectangle, minimum width=1cm, minimum height=0.6cm, text width=1cm, align=center},
                correct/.style={box, fill=green!20},
                error/.style={box, fill=red!20}
            ]
            \matrix (conmat) [row sep=.05cm, column sep=.05cm] {
                \node (tp) [correct] {1372}; &
                \node (fn) [error] {575}; \\
                \node (fp) [error] {224}; &
                \node (tn) [correct] {6515}; \\
            };
            \node [left=.2cm of conmat, yshift=0.7cm, rotate=90, font=\bfseries] {Actual};
            \node [above=.05cm of conmat, font=\bfseries] {Predicted};
            \end{tikzpicture}
            {\small HB Eval Probe}
        \end{subfigure}
        \hspace{1.5cm}
        \begin{subfigure}{0.25\linewidth}
            \centering
            \begin{tikzpicture}[
                box/.style={draw, rectangle, minimum width=1cm, minimum height=0.6cm, text width=1cm, align=center},
                correct/.style={box, fill=green!20},
                error/.style={box, fill=red!20}
            ]
            \matrix (conmat) [row sep=.05cm, column sep=.05cm] {
                \node (tp) [correct] {225}; &
                \node (fn) [error] {78}; \\
                \node (fp) [error] {166}; &
                \node (tn) [correct] {1773}; \\
            };
            \node [left=.2cm of conmat, yshift=0.7cm, rotate=90, font=\bfseries] {Actual};
            \node [above=.05cm of conmat, font=\bfseries] {Predicted};
            \end{tikzpicture}
            {\small HB Hard Probe}
        \end{subfigure}
        \caption{MedLLaMA Model}
        \label{fig:medllama_cm}
    \end{subfigure}
    \caption{Confusion matrices for rubric-only judge-label classification across the four evaluated models}
    \label{fig:all_cm}
\end{figure*}

\begin{figure*}[!b]
    \centering
    \includegraphics[width=0.8\textwidth]{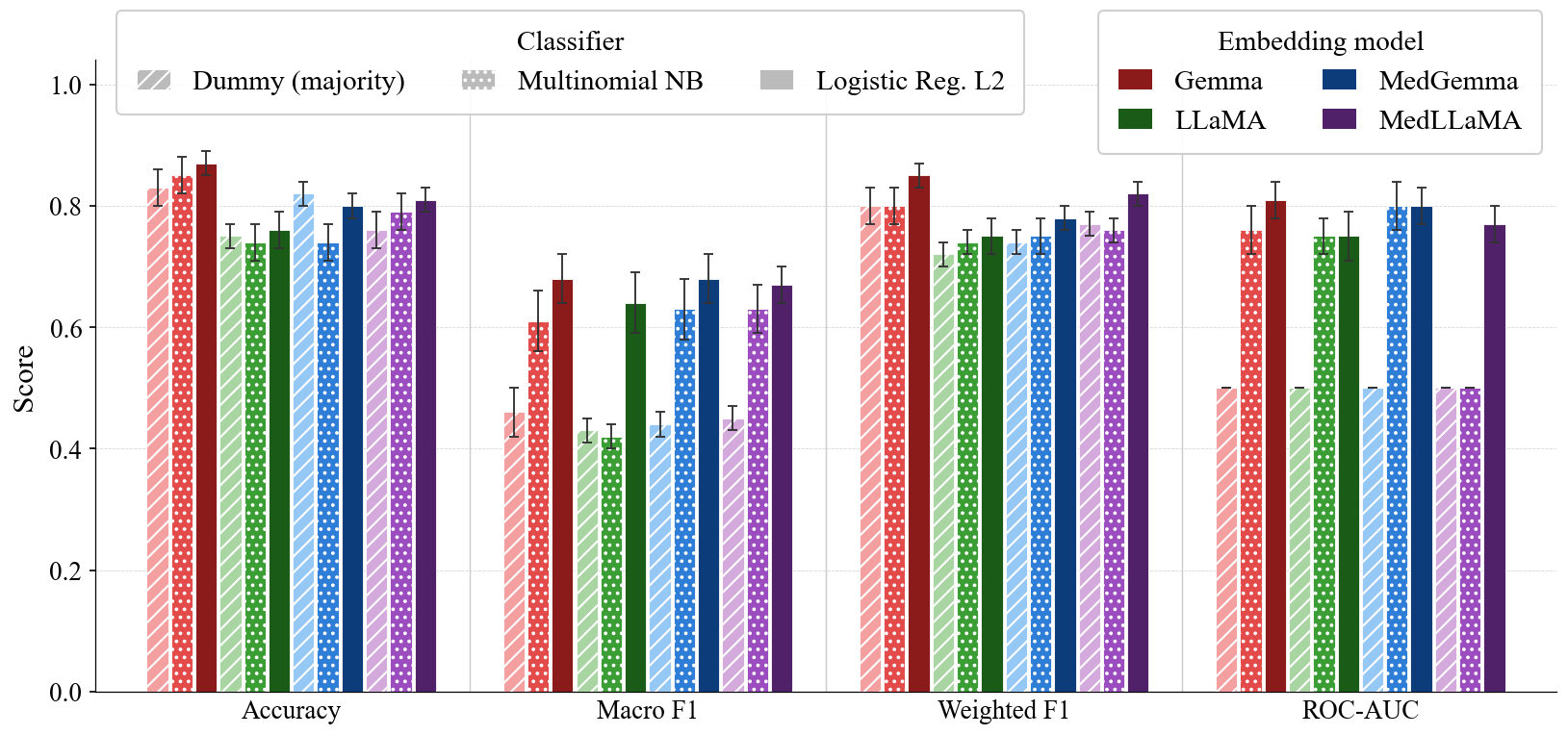}
    \caption{Rubric-Only Classifiers Achieve Nontrivial Performance across Judge Models and Evaluation Metrics}
    \label{fig:classifier_performance_summary}
\end{figure*}

\noindent Figure~\ref{fig:classifier_performance_summary} reports cross-validated accuracy, macro-F1, weighted-F1, and ROC-AUC across four embedding models used as judges:  Gemma, LLaMA, MedGemma, and MedLLaMA. Across all four metrics, logistic regression consistently outperforms both the Naïve Bayes baseline and the majority-class dummy, confirming that rubric-derived lexical features carry a discriminative signal beyond class imbalance. This pattern holds across embedding families, though performance varies by judge, suggesting that different models impose distinct evaluative surfaces on the same rubric text. These results strengthen the case that rubric-based LLM evaluation pipelines warrant closer scrutiny.

\section{Validation and Uncertainty Estimation}
\label{app:cross_val_proto}

\subsection{Classifier Evaluation Protocol}
\label{app:classifier-validation}
To assess the robustness of the rubric-only probe results, we use both cross-validation and held-out test-set evaluation. For each probe cohort, we first reserve 20\% of the data as a held-out test set. The remaining 80\% is used for model development and evaluated using standard 5-fold cross-validation. In each fold, four partitions are used for training, and the remaining partition is used for validation, with each partition serving as the validation set exactly once. We report the mean and standard deviation across the five folds.

\noindent For the final held-out evaluation, the classifier is evaluated on the reserved test set, which is not used during model development. This provides an independent estimate of performance in addition to the cross-validation results. Detailed results for the four judge cohorts are reported in Table~\ref{tab:cv_bootstrap_results}.

\subsection{Bootstrap Confidence Intervals}
\label{app:bootstrap}
To quantify sampling uncertainty in the held-out evaluation, we perform non-parametric bootstrap resampling with 10,000 iterations over the held-out test set. For each bootstrap sample, we recompute the evaluation metrics and use the resulting empirical distributions to obtain 95\% confidence intervals.

\noindent Across the four judge cohorts, the resulting confidence intervals are relatively narrow, indicating that the observed probe performance is stable with respect to sampling variability. We report confidence intervals for accuracy, precision, recall, F1, and balanced accuracy in Table~\ref{tab:cv_bootstrap_results}.

\begin{table*}[t]
\centering
\small
\begin{tabular}{llcc}
\toprule
\textbf{Model} & \textbf{Metric} &
\textbf{5-Fold CV (Mean $\pm$ SD)} &
\textbf{Held-Out Test [95\% CI]} \\
\midrule
\multirow{5}{*}{Gemma}
& Accuracy & 0.826 $\pm$ 0.006 & 0.826 [0.814, 0.837] \\
& F1 & 0.682 $\pm$ 0.006 & 0.682 [0.671, 0.693] \\
& Precision & 0.687 $\pm$ 0.016 & 0.687 [0.656, 0.717] \\
& Recall & 0.704 $\pm$ 0.013 & 0.704 [0.678, 0.729] \\
& Balanced Acc. & 0.801 $\pm$ 0.006 & 0.801 [0.789, 0.812] \\
\midrule
\multirow{5}{*}{MedGemma}
& Accuracy & 0.886 $\pm$ 0.004 & 0.886 [0.879, 0.893] \\
& F1 & 0.781 $\pm$ 0.004 & 0.781 [0.773, 0.789] \\
& Precision & 0.772 $\pm$ 0.011 & 0.772 [0.751, 0.793] \\
& Recall & 0.785 $\pm$ 0.009 & 0.785 [0.768, 0.801] \\
& Balanced Acc. & 0.867 $\pm$ 0.004 & 0.867 [0.860, 0.873] \\
\midrule
\multirow{5}{*}{LLaMA}
& Accuracy & 0.842 $\pm$ 0.005 & 0.842 [0.833, 0.851] \\
& F1 & 0.705 $\pm$ 0.004 & 0.705 [0.697, 0.713] \\
& Precision & 0.707 $\pm$ 0.010 & 0.707 [0.688, 0.726] \\
& Recall & 0.721 $\pm$ 0.011 & 0.721 [0.700, 0.741] \\
& Balanced Acc. & 0.817 $\pm$ 0.004 & 0.817 [0.809, 0.825] \\
\midrule
\multirow{5}{*}{MedLLaMA}
& Accuracy & \textbf{0.900} $\pm$ 0.003 & 0.900 [0.894, 0.906] \\
& F1 & \textbf{0.800} $\pm$ 0.003 & 0.800 [0.794, 0.806] \\
& Precision & \textbf{0.798} $\pm$ 0.008 & 0.798 [0.783, 0.812] \\
& Recall & \textbf{0.804} $\pm$ 0.007 & 0.804 [0.791, 0.816] \\
& Balanced Acc. & \textbf{0.881} $\pm$ 0.003 & 0.881 [0.876, 0.886] \\
\bottomrule
\end{tabular}
\caption{Five-fold cross-validation and held-out test-set performance of the rubric-only probe classifier. Confidence intervals are obtained using 10,000 bootstrap resamples of the held-out test set.}
\label{tab:cv_bootstrap_results}
\end{table*}

\subsection{Cross-Dataset Generalization}
\label{app:cross-dataset}
To further assess the generalizability of the rubric-only predictive signal, we conduct a cross-dataset evaluation between \textsc{HealthBench-Eval-Probe} and \textsc{HealthBench-Hard-Probe}. Specifically, we train the rubric classifier on one probe dataset and evaluate it on the other, considering both transfer directions: training on \textsc{HealthBench-Eval-Probe} and testing on \textsc{HealthBench-Hard-Probe}, and vice versa.

\begin{figure}[!b]
    \centering
    \includegraphics[width=0.8\linewidth]{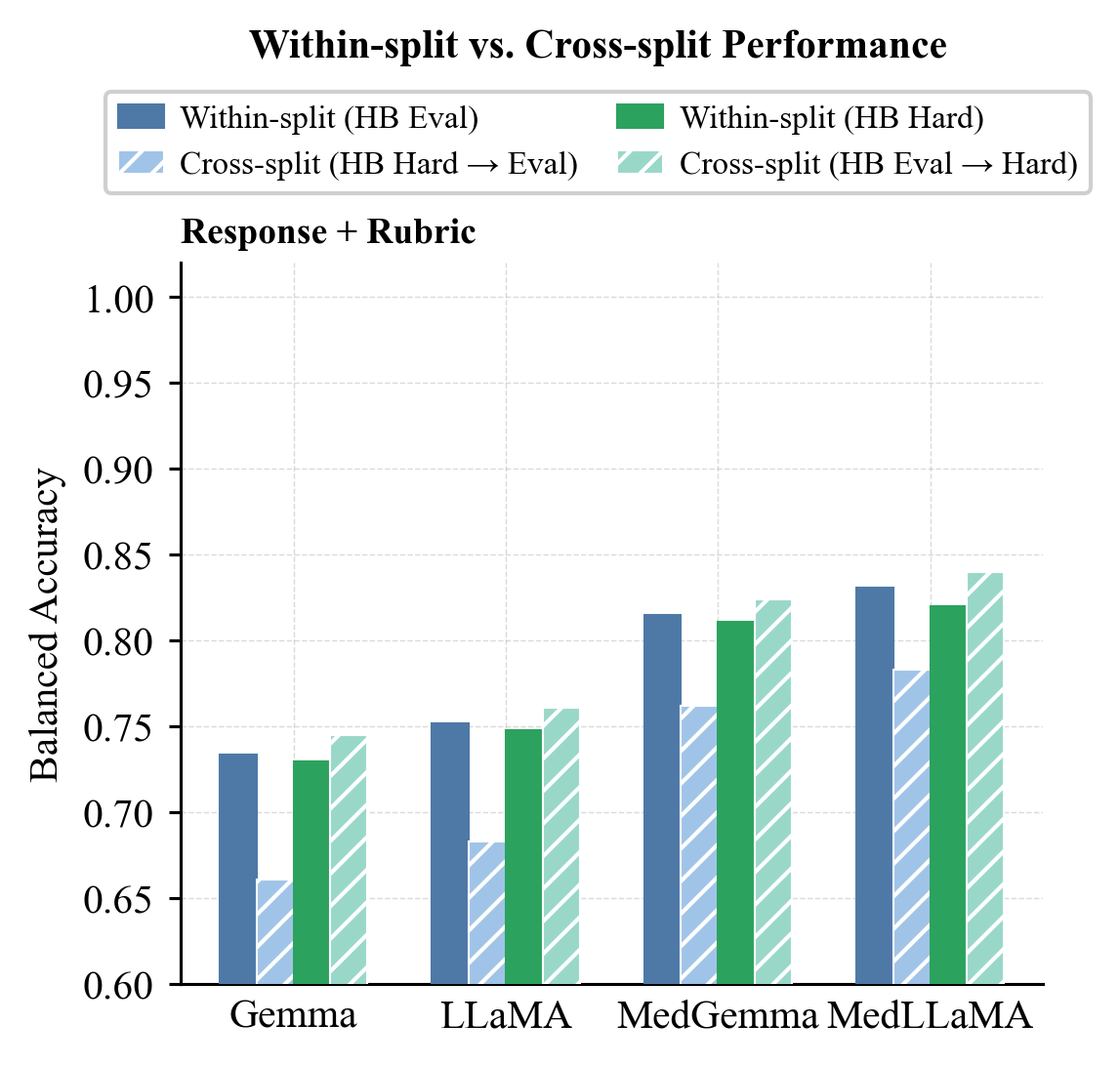}
    \caption{Cross-dataset transfer: within-split vs. cross-split balanced accuracy.
    Solid bars = within-split (train = test). Hatched bars = cross-split (train on one split, test on the other)
    Lower gap = better generalization.}
    \label{fig:cross}
\end{figure}

\noindent This setting provides a more stringent test of generalization because the classifier is evaluated on rubrics from a different \textsc{HealthBench} variant than those used during training. Figure~\ref{fig:cross} presents the resulting accuracy across the two transfer directions. Performance remains above the 0.5 chance level under both cross-dataset settings, indicating that the rubric-only classifier captures predictive patterns that transfer across the two \textsc{HealthBench} variants rather than being entirely specific to a single dataset.

\noindent The cross-dataset results therefore provide an additional check that the observed rubric--label association is not solely an artifact of the particular train--test partition or dataset-specific distribution. Detailed results across the different judge cohorts are reported in Figure~\ref{fig:cross}.

\section{Classifier Ablations and Baselines}
\label{app:classifier-ablations}

\subsection{Choice of PubMedBERT}
\label{app:pubmedbert}
We use PubMedBERT as the primary probe classifier because the \textsc{HealthBench} rubrics contain specialized biomedical terminology, for which domain-specific pretraining provides a natural advantage. To assess whether our findings depend on this architectural choice, we additionally evaluate BERT, RoBERTa, and DeBERTa-v3 under the same experimental setting.

\noindent Table~\ref{tab:classifier_comparison} reports the resulting performance across the three judge cohorts. All four classifiers exhibit the same qualitative pattern, with performance substantially above chance. PubMedBERT provides a modest, generally consistent improvement, supporting its use as the primary probe and indicating that the observed rubric-only signal is not specific to a single classifier architecture.

\begin{table*}[t]
\centering
\small
\begin{tabular}{lccc}
\toprule
\textbf{Classifier} &
\textbf{Gemma} &
\textbf{MedGemma} &
\textbf{LLaMA} \\
& Acc / Pre / Rec / F1 &
Acc / Pre / Rec / F1 &
Acc / Pre / Rec / F1 \\
\midrule
PubMedBERT
& \textbf{86.5} / 83.1 / 86.2 / \textbf{85.9}
& \textbf{79.2} / 82.1 / \textbf{79.1} / 80.6
& \textbf{78.2} / 77.1 / \textbf{78.0} / \textbf{77.8} \\
BERT
& 86.1 / \textbf{85.2} / 86.2 / 84.6
& 76.0 / 80.9 / 76.1 / 78.4
& 74.5 / 78.2 / 74.1 / 75.9 \\
RoBERTa
& 85.2 / 81.5 / \textbf{90.2} / 85.3
& 78.4 / \textbf{83.7} / 78.2 / \textbf{81.1}
& 75.9 / \textbf{78.7} / 75.4 / 76.6 \\
DeBERTa-v3
& 84.6 / 84.5 / 86.7 / 84.9
& 70.5 / 83.4 / 71.6 / 74.3
& 74.2 / 78.3 / 74.7 / 75.1 \\
\bottomrule
\end{tabular}
\caption{Classifier comparison on \textsc{HealthBench}. Values are reported
as accuracy, precision, recall, and F1 (\%).}
\label{tab:classifier_comparison}
\end{table*}

\begin{table}[!b]
\centering
\small
\begin{tabular}{lcc}
\toprule
\textbf{Response Model} &
\textbf{\textsc{HB}-Eval} &
\textbf{\textsc{HB}-Hard} \\
& $n=4000$ & $n=1000$ \\
\midrule
Gemma & 89.4 & 76.0 \\
LLaMA & 88.1 & 79.3 \\
MedGemma & 83.4 & 78.3 \\
MedLLaMA & 78.3 & 88.4 \\
\bottomrule
\end{tabular}
\caption{Majority-class accuracy (\%) for each model on
\textsc{HealthBench}-Eval \textsc{HealthBench}-Hard.}
\label{tab:majority-baseline}
\end{table}

\subsection{Majority-Class Baseline}
\label{app:majority-baseline}
Because each response model produces different candidate responses for the same rubrics, the resulting LLM-as-a-Judge labels have different class distributions. We therefore compute the majority-class baseline separately for each response model by always predicting its dominant label (Table ~\ref{tab:majority-baseline}).

\noindent Although the majority baseline can achieve high accuracy under class imbalance, it obtains a balanced accuracy of 0.50 by construction. We therefore use weighted classification and balanced subsampling in the main experiments to guard against a trivial majority-class solution. The resulting probe performance remains above this baseline across metrics.

\section{Counterfactual Perturbations}
\label{app:counterfactual}
This section provides additional details on the counterfactual experiments presented in Section~\ref{sec:counterfactual}.

\subsection{Output Perturbation}
\label{app:output-perturbation}
For each conversation--rubric pair $(C_i, r_i)$, Mistral-7B was prompted to generate two responses: one designed to satisfy the rubric ($R_{\mathrm{satisfy}}$) and one designed to violate it ($R_{\mathrm{violate}}$). We manually inspected the generated pairs to verify that the intended contrast was preserved. The resulting responses were then evaluated independently by Qwen2.5-7B-Instruct using the original conversation
and rubric. 

\noindent We compare the judge's verdicts across each response pair (Table~\ref{tab:output-counterfactual}). A \textbf{desired flip} occurs when the judge changes its verdict in accordance with the intended change in rubric satisfaction. A \textbf{sticky} outcome occurs when the judge retains its original verdict despite the response perturbation. A \textbf{wrong-direction flip} occurs when the judge changes its verdict in the opposite direction.

\begin{table}[h]
\centering
\small
\begin{tabular}{lcc}
\toprule
\textbf{Outcome} & \textbf{Count} & \textbf{Percentage} \\
\midrule
Desired flip & 186 & 37.7\% \\
Wrong-direction flip & 88 & 17.8\% \\
Sticky satisfy & 107 & 21.7\% \\
Sticky violate & 112 & 22.7\% \\
\bottomrule
\end{tabular}
\caption{Transition outcomes under counterfactual output perturbation. Percentages are computed over the corresponding 493 valid paired evaluations.}
\label{tab:output-counterfactual}
\end{table}

\noindent We summarize the paired outcomes using three measures:
\begin{equation}
\small{\text{Desired Flip Rate}
=
\frac{\text{Desired Flips}}{\text{Valid Paired Perturbations}}},
\end{equation}
\begin{equation}
\small{\text{Stickiness Rate}
=
\frac{\text{Sticky Satisfy}+\text{Sticky Violate}}
{\text{Valid Paired Perturbations}}},
\end{equation}
\begin{equation}
\small{\text{Output Inconsistency Rate}
=
1-\text{Desired Flip Rate}}.
\end{equation}

\noindent For the 493 valid paired perturbations, the desired-flip rate is 37.7\%, the stickiness rate is 39.6\%, and the resulting inconsistency rate is 62.3\%. These results provide the detailed transition statistics underlying the complete analysis.

\subsection{Rubric Perturbation}
\label{app:rubric-perturbation}
We randomly sampled 1,000 \textsc{HealthBench} conversations together with candidate responses generated by Gemma and LLaMA. For each instance, Mistral-7B was used to construct a counterfactual rubric whose intended criterion was reversed relative to the original rubric. The conversation and candidate response were kept fixed. We manually inspected the resulting rubric pairs to verify that the intended semantic reversal was preserved.

\begin{table}[b]
\centering
\small
\begin{tabular}{lcc}
\toprule
\textbf{Outcome} & \textbf{Gemma} & \textbf{LLaMA} \\
\midrule
Desired flip & 125 + 43 (16.8\%) & 252 + 70 (32.2\%) \\
Sticky satisfy & 8 (0.8\%) & 31 (3.1\%) \\
Sticky violate & 824 (82.4\%) & 646 (64.7\%) \\
\bottomrule
\end{tabular}
\caption{Transition outcomes under counterfactual rubric perturbation. The two desired flip directions are aggregated in the reported desired flip rate.}
\label{tab:rubric-counterfactual}
\end{table}

\noindent We compare the judge's verdicts before and after the rubric perturbation (Table~\ref{tab:rubric-counterfactual}). A \textbf{desired flip} occurs when the judge reverses its verdict in accordance with the reversed rubric criterion, while a \textbf{sticky} outcome occurs when the judge retains its original verdict despite the change in rubric semantics or phrasing.

\noindent Aggregating the two desired-flip directions, the judge produces the expected reversal in 16.8\% of Gemma pairs and 32.2\% of LLaMA pairs. The corresponding rubric inconsistency rates are 83.2\% and 67.8\%, respectively. Thus, the detailed transition analysis confirms the substantial non-responsiveness reported above.

\section{Rubric Semantics Exploration}
\label{app:semantics}
\subsection{Keywords}
To examine whether rubric text encodes lexically recoverable evaluative priors, we conduct two complementary analyses at the word level: binary lexical cue profiling across judge labels, and contrastive token attribution stratified by class.

\begin{figure}[t]
    \centering
    \includegraphics[width=\linewidth]{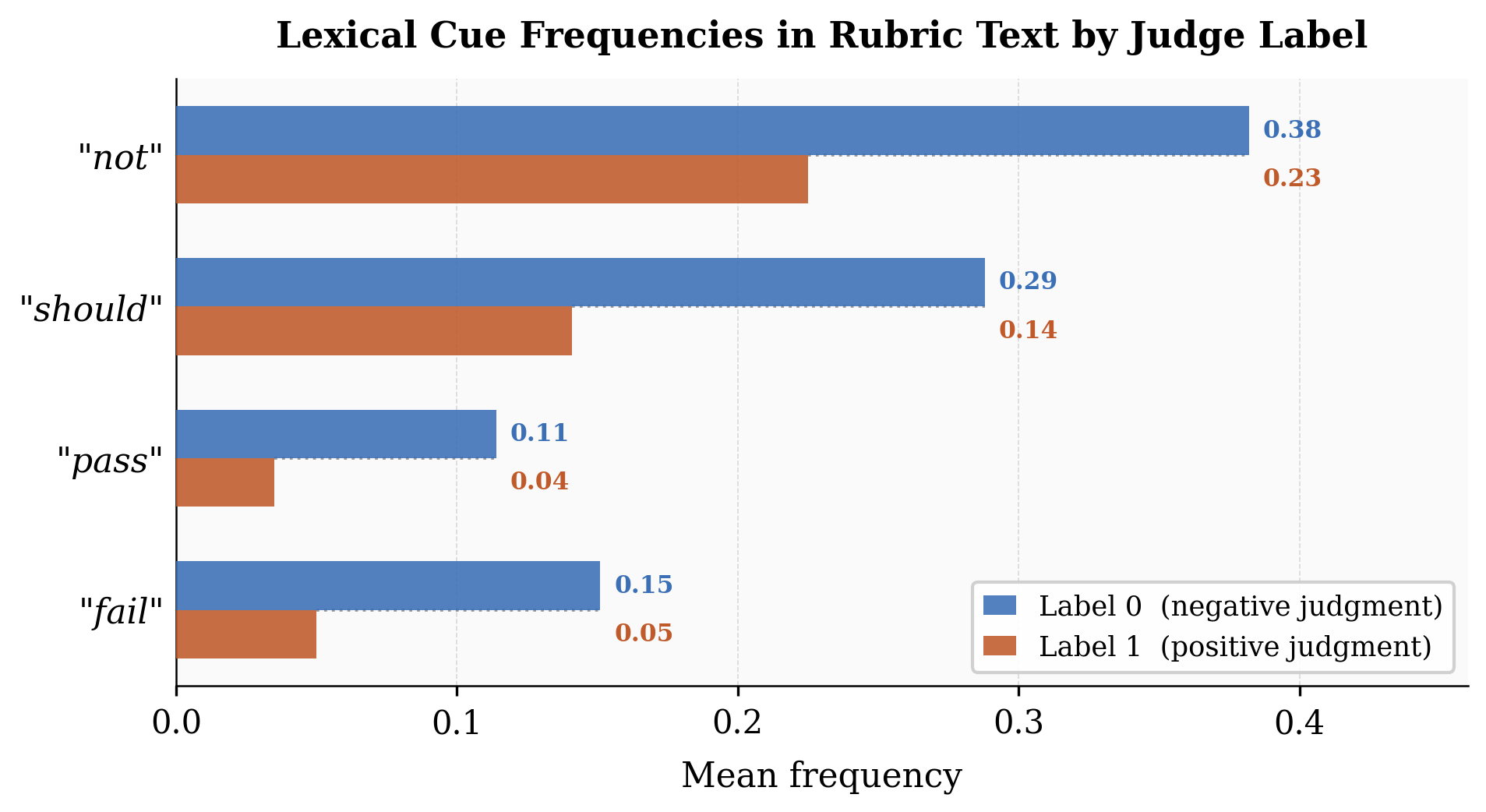}
    \caption{Lexical Cue Frequencies in Rubric Text.\\
    Mean frequency of selected binary lexical features across rubrics stratified by judge label.}
    \label{fig:lexical_cues}
\end{figure}

\noindent Figure~\ref{fig:lexical_cues} shows that several surface tokens exhibit pronounced frequency asymmetries across judge labels. The terms fail, pass, and not appear substantially more often in rubrics assigned label 0, while label 1 rubrics show comparatively sparse usage of these markers. The word \emph{should} is elevated in both classes but skews toward label 0, suggesting prescriptive and negation-heavy language is a consistent surface correlate of negative judgments.

\begin{figure}[!b]
    \centering
    \includegraphics[width=\linewidth]{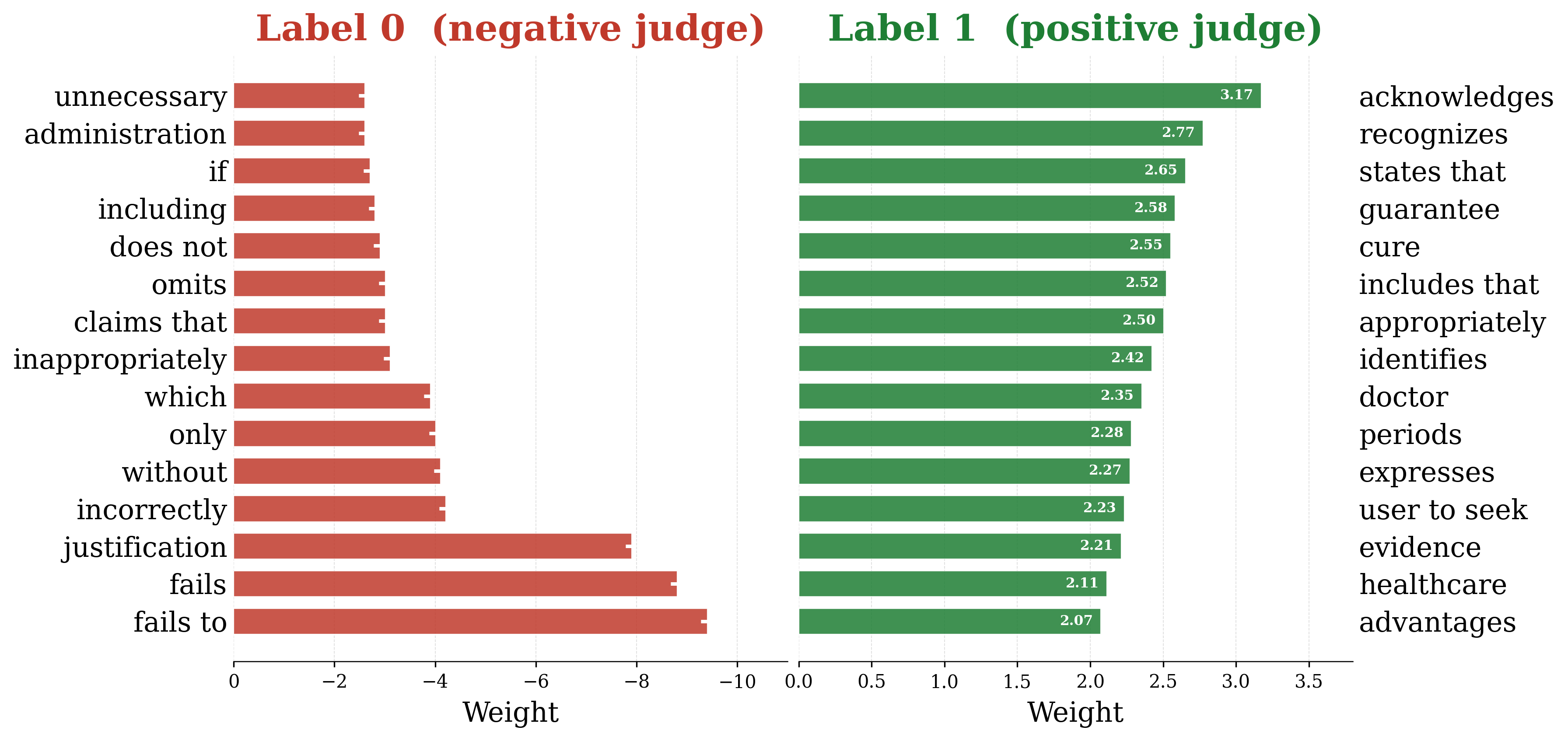}
    \caption{Top Logistic Regression Weights by Judge Label. Rubric phrasing encodes a class-discriminative lexical signal}
    \label{fig:token_attribution}
\end{figure}

\noindent Figure~\ref{fig:token_attribution} presents the top logistic regression TF-IDF weights for each class, displayed as a diverging chart. Rubrics associated with negative judgments (label 0) are dominated by failure-marking and negation vocabulary, with justification also carrying strong negative weight, suggesting that rubrics that foreground what must be justified tend to predict failure. Positive-judgment rubrics (label 1) show the mirror pattern: the highest-weighted tokens are epistemic and affirmative verbs, reflecting rubric language that describes what a correct response should exhibit rather than what it must not do. The two vocabularies are entirely disjoint, indicating that failure-predictive language is encoded more strongly and distinctively in rubric text than success-predictive language.

\subsection{Word Length Analysis}
\begin{figure}[t]
    \centering
    \includegraphics[width=0.9\linewidth]{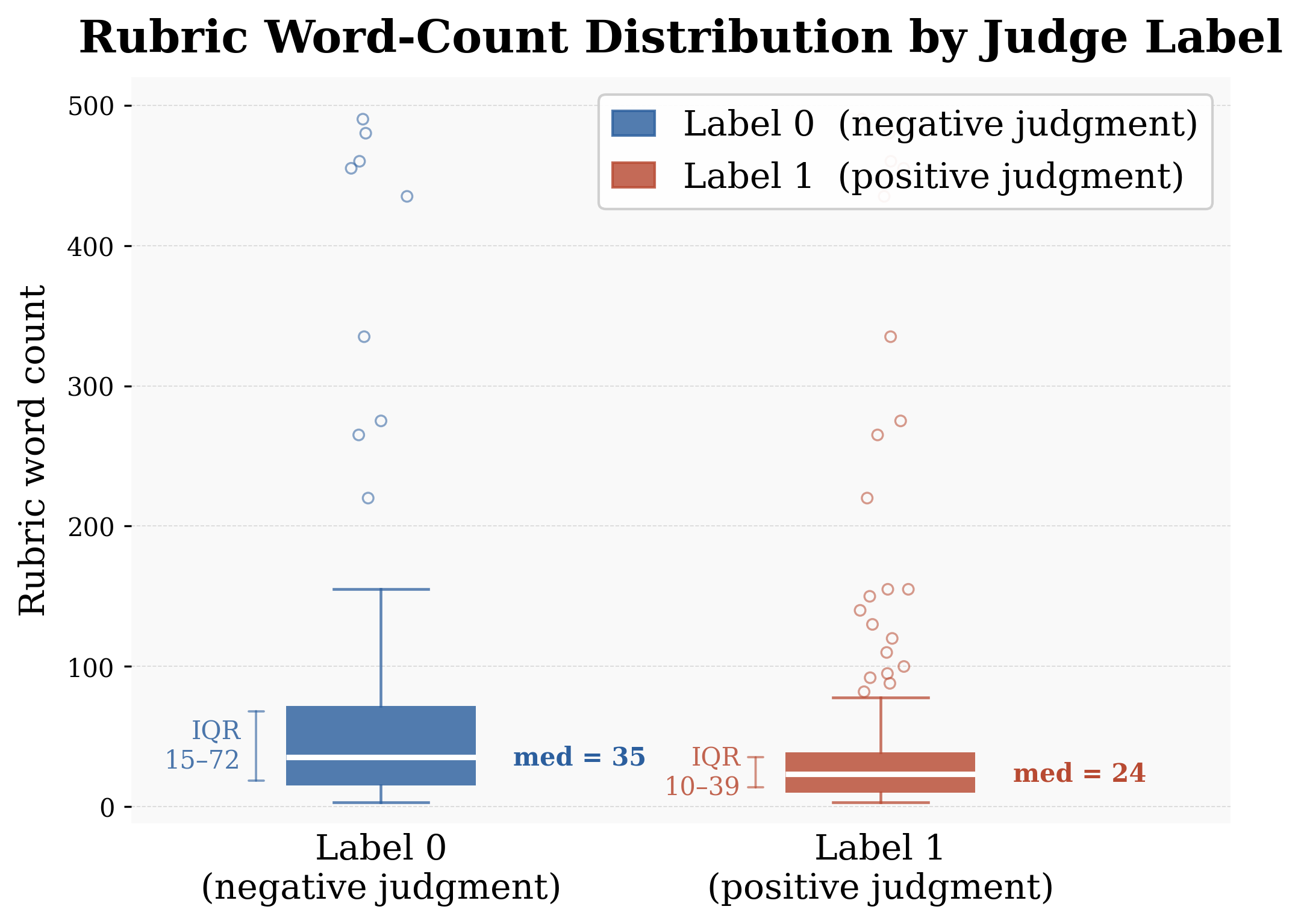}
    \caption{Rubric Word-Count Distribution by Judge Label. Label 0 rubrics are longer on average; distributions overlap substantially.}
    \label{fig:word-len}
\end{figure}

Figure~\ref{fig:word-len} shows the distribution of rubric word counts stratified by judge label. Label 0 rubrics are systematically longer than label 1 rubrics. Both the distributions overlap substantially, so word count alone does not cleanly separate the two classes. The many outliers in both labels suggest that a small number of very long rubrics exist in each group.
\subsection{BERTopic Modeling}
To move beyond individual tokens and examine the sentence-level semantic structure of rubric text, we apply BERTopic~\cite{grootendorst2022bertopic} conditioned on the judge label. Rather than jointly discovering topics across all rubrics, we fit separate topic models for each class, allowing us to characterize the evaluative concerns prototypical of positive and negative judgments independently.
\begin{figure}[!b]
    \centering    \includegraphics[trim={0 2cm 0 0},clip,width=0.9\columnwidth]{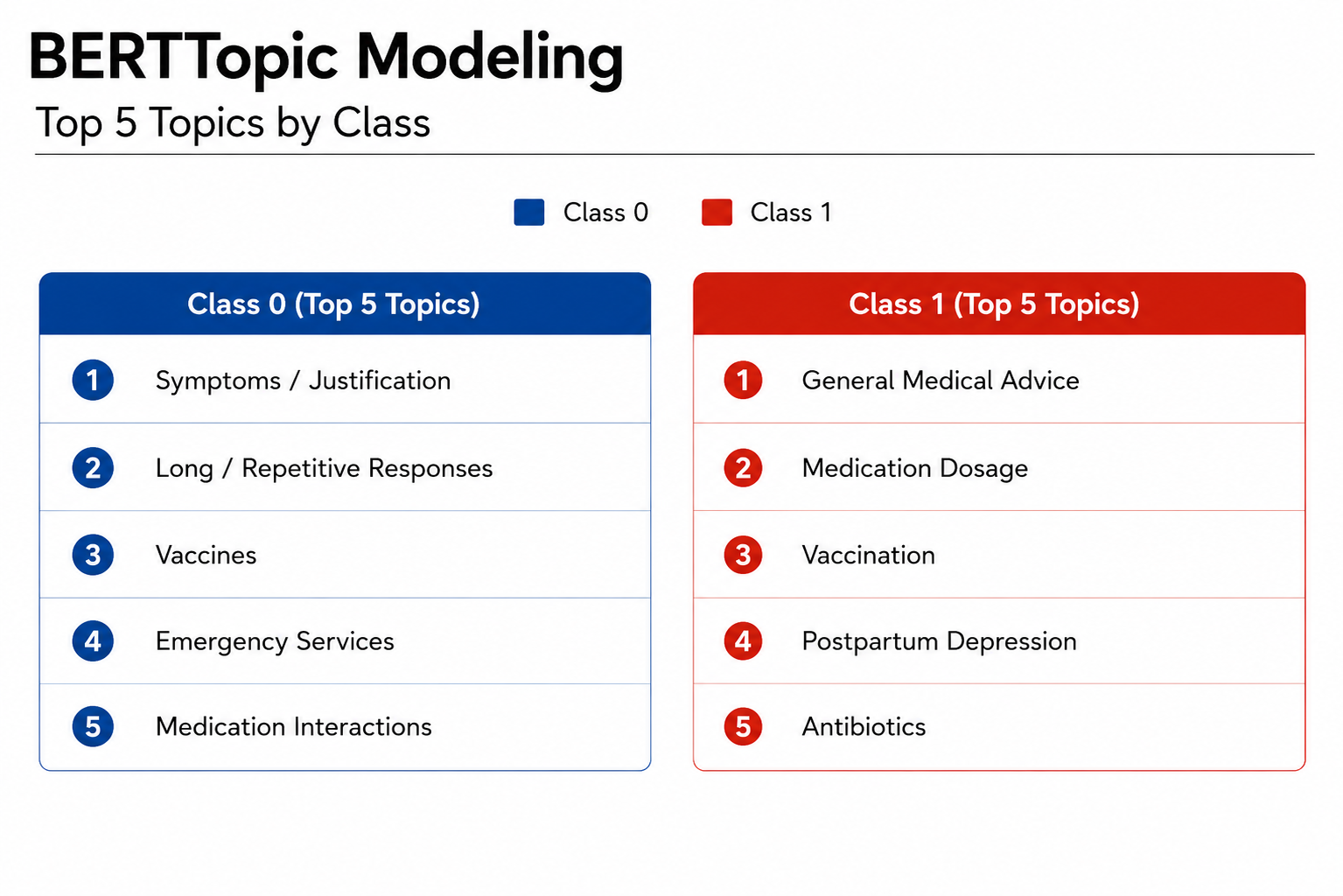}
    \caption{BERTopic Model results on \textsc{HealthBench-Hard} and \textsc{HealthBench-Eval}}
    \label{fig:bertopic}
\end{figure}
\\
The top five topics per class, shown in Figure~\ref{fig:bertopic}, reveal a semantically coherent contrast. Class 0 rubrics cluster around failure modes and clinical edge cases (symptom justification requirements, verbose or repetitive responses, and scenarios involving vaccines, emergency services, and medication interactions) consistent with the negation-heavy lexical profile identified earlier. Class 1 rubrics organize around affirmative clinical competencies: general medical advice, dosage correctness, vaccination guidance, postpartum mental health, and antibiotic stewardship.\\
The two topic sets are largely disjoint, with no direct thematic overlap in the top five, confirming that positive and negative judgment rubrics occupy distinct semantic regions. That these regions correspond to coherent clinical subfields rather than generic discourse patterns demonstrates that there is evaluative signal in rubric text.

\end{document}